\documentclass{article}
\PassOptionsToPackage{numbers, compress}{natbib}

\usepackage[final]{neurips_2022}

\usepackage[utf8]{inputenc} 
\usepackage[T1]{fontenc}    
\usepackage{hyperref}       
\usepackage{url}            
\usepackage{booktabs}       
\usepackage{amsfonts}       
\usepackage{nicefrac}       
\usepackage{microtype}      
\usepackage[pdftex]{graphicx}
\graphicspath{ {./Figures/} }

\title{Learning Invariant World State Representations with Predictive Coding}

\author{%
  Avi Ziskind \\
  SRI International\\
  \texttt{avi.ziskind@sri.com} \\

  \And
  Sujeong Kim \\
  SRI International \\
  \texttt{sujeong.kim@sri.com} \\
  
  \AND
  Giedrius T. Burachas \\
  SRI International \\
  \texttt{giedrius.burachas@sri.com} \\

}

\begin{document}

\maketitle

\begin{abstract}
Self-supervised learning methods overcome the key bottleneck for building more capable AI: limited availability of labeled data. However, one of the drawbacks of self-supervised architectures is that the representations that they learn are implicit and it is hard to extract meaningful information about the encoded world states, such as 3D structure of the visual scene encoded in a depth map. Moreover, in the visual domain such representations only rarely undergo evaluations that may be critical for downstream tasks, such as vision for autonomous cars. Herein, we propose a framework for evaluating visual representations for illumination invariance in the context of depth perception. We develop a new predictive coding-based architecture and a hybrid fully-supervised/self-supervised learning method. We propose a novel architecture that extends the predictive coding approach: PRedictive Lateral bottom-Up and top-Down Encoder-decoder Network (PreludeNet), which explicitly learns to infer and predict depth from video frames. In PreludeNet, the encoder’s stack of predictive coding layers is trained in a self-supervised manner, while the predictive decoder is trained in a supervised manner to infer or predict the depth. 
We evaluate the robustness of our model on a new synthetic dataset, in which lighting conditions (such as overall illumination, and effect of shadows) can be be parametrically adjusted while keeping all other aspects of the world constant.
PreludeNet achieves both competitive depth inference performance and next frame prediction accuracy. We also show how this new network architecture, coupled with the hybrid fully-supervised/self-supervised learning method, achieves balance between the said performance and invariance to changes in lighting. 
The proposed framework for evaluating visual representations can be extended to diverse task domains and invariance tests.

\end{abstract}

\section{Introduction}

Deep learning has made much progress in image and video understanding. Even in the age of emerging generalist approaches (\cite{alayrac2022flamingo,radford2021learning}), though, most vision tasks require custom network architectures, which often do not generalize beyond the specific domain (e.g. depth, optical flow, object detection etc.). Even domain-specific models often perform only within a narrow range of relevant parameters such as illumination, and fail in the edge cases of poor illumination or extreme shadows. 

Developing AI approaches for fully autonomous scene understanding requires developing richer architectures that can scale from simple image/pixel-based predictions (e.g. depth, optical flow) to multiple tasks at higher semantic level (navigation, path planning, obstacle avoidance) and that can handle unexpected domain shifts. Robust autonomous navigation algorithms should also be able predict future states of the world and develop plans according to those predictions, but also be quick to adapt and be flexible if expectations are violated (e.g. cars changing lanes without warning, a deer wanders into the lane etc.). We propose that incorporating predictive coding motifs in deep learning networks is a promising approach to developing a generic vision system for image and video scene understanding capable of supporting multiple computer vision tasks in an invariant and novelty-aware manner. We have developed a predictive-coding-based deep learning approach that can be used to predict depth, optical flow, self-motion, and semantics from video frames, using a unified architecture. In this paper we evaluate our approach in a set of experiments addressing next-frame video prediction and depth estimation accuracy as well as invariance of the model's inferences to variations in illumination.\footnote{Code for training and running our code will be made publicly available upon publication}

\begin{figure}
 \centering
 \includegraphics[scale=0.68]{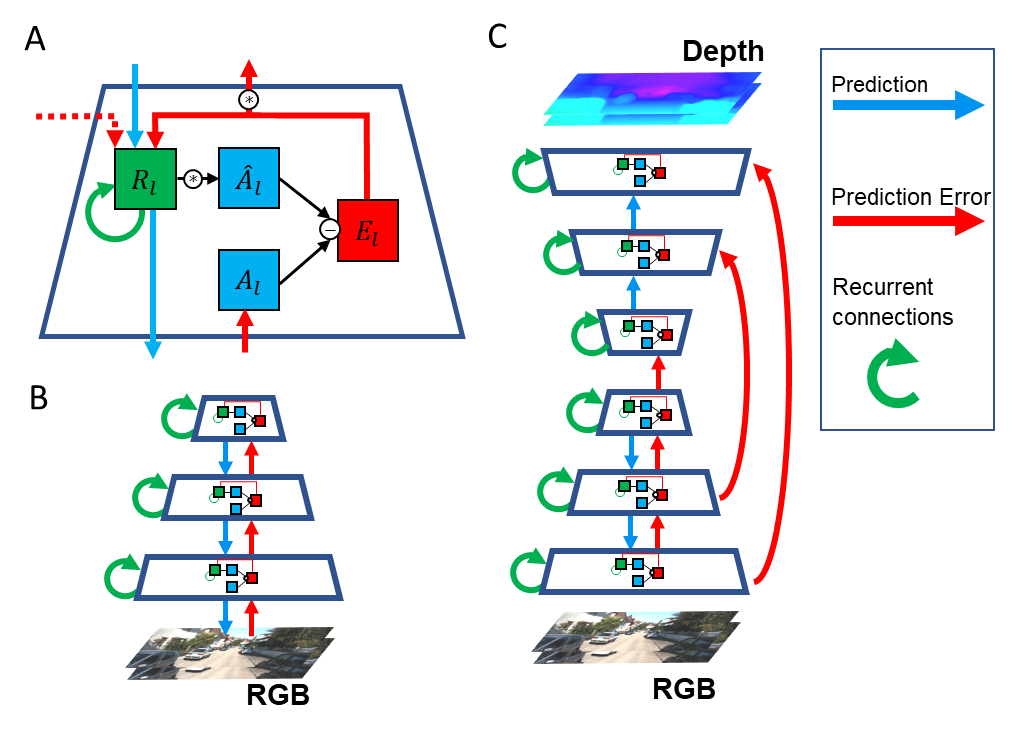}
 \caption{Predictive Coding architecture. \textbf{A}. Predictive coding block (adapted from \cite{lotter_deep_2017}). In each block, a Recurrent unit $R$ makes a prediction ($\hat{A}_l$) of what the input to the layer ($A_l$) will be. The prediction error unit 
 $E_l$ is the difference between the predicted and actual inputs, which are fed back to the recurrent unit and as inputs the subsequent layer. Circles with a * indicate standard convolutional layers. \textbf{B}. PredNet Architecture, consisting of a stack of three predictive coding blocks. In each time-step, first the predictions are propagated top-down (blue arrows). This is followed by a sequence of bottom-up prediction error propagation steps (red arrows). C: PreludeNet encoder/decoder architecture, consisting of three RGB-encoding blocks, followed by three Depth-decoding blocks, Encoder and decoder blocks are connected with lateral connections (red arcs).}
 \label{fig:prednet}
\end{figure}

\paragraph{Predictive Coding.}
Hierarchical predictive coding \cite{rao_predictive_1999} has been introduced as a generative model for addressing learning and inference in the brain. It features an efficient neural coding scheme for learning implicit world state representations by minimizing the predicted future states error. Neuroscientific studies have shown that predictive coding is a recurring computational motif across brain regions \cite{rao_predictive_1999, whittington_theories_2019}, and can explain a plethora of neuronal response properties in visual, auditory, and cognitive control regions of the brain.

The core idea of predictive coding theory when applied to sensory processing is that individual computational units in the brain learn an internal representation to predict future feed-forward inputs from earlier sensory regions. A deviation of the actual inputs from their expected inputs gives rise to a prediction "error" signal, which is used to update the local representation of the computational unit, and is also passed up as a signal to higher brain regions, where this cycle repeats. There are several computational advantages to this approach: (1) Efficiency: since only prediction error needs to be propagated, and not a representation of the entire signal, the neurons can apply their entire dynamic range to just the error representation. (2) Saliency: it provides a natural way for new, salient inputs to quickly propagate along the sensory pathway. Unexpected inputs (i.e. inputs that have a large deviation from their expected values) generate large prediction error signals, which quickly propagate in a bottom up fashion. (3) Hierarchy: Predictive coding can be extended to multiple time-scales across different representation hierarchies resulting in learning of representations for long term predictions. Learning can be dynamic and adaptive, continuously being driven by what is not already captured by current representation. 
Predictive coding has recently been used with self-supervised deep learning frameworks (e.g. \cite{lotter_deep_2017}; see Fig. \ref{fig:prednet} A,B), and resulted in state-of-the-art video frame prediction. Note that in the this paper we are not addressing distant cousins of the neuro-inspired approach, such as contrastive predictive coding \cite{oord2018representation}.  We use the advantages of predictive coding to develop a hybrid fully- and self-supervised version of predictive coding for next frame and depth prediction.

Our contributions include:

\begin{itemize}
 \item Illustrating how to extract latent information learned by a predictive coding algorithm, using predictive coding principles and hybrid learning.
 \item Showing how the latent information required for next-frame prediction can be used to generate depth maps
 \item A new rendering simulator and dataset (SINAV) to test the robustness to changes in illumination of our and other approaches to depth-prediction.
 
\end{itemize}

\section{Approach and related work}

Our approach builds on the deep learning network implementation of the predictive coding principle for video frame prediction, PredNet \cite{lotter_deep_2017}. Despite the computational advantages to predictive coding, its use in deep learning applications has been limited.

\paragraph{PredNet for next-frame prediction.}
\citet{lotter_deep_2017} showed that the predictive-coding computational motif can be efficiently implemented a deep neural network, which they called PredNet. The PredNet model implements the above-mentioned predictive coding motif as shown in Fig. \ref{fig:prednet}A. In each layer $l$, there is a recurrent unit, $R_l$, implemented as a convolutional LSTM \cite{shi_convolutional_2015}, which makes predictions about the feed-forward input that it will receive from layer $l - 1$ (for the bottom most unit, $l=0$, the input is the raw image pixels). The predicted input $\hat{A}_l$, is compared with the actual input, $A_l$, and the prediction error for that layer, $E_l$ is calculated as the difference between them: $E_l=\hat{A}_l-A_l$. (This predictive error signal is actually the concatenation of the positive and negative components of the error: $E_l = Concat[ReLU(A_l-\hat{A}_l), ReLU(\hat{A}_l-A_l)]$). This error signal feeds back into the recurrent unit $R_l$ to update the recurrent state, and also becomes (after a convolutional layer) the input to the next layer, $A_{l+1}$, (unless layer $l$ is the top-most layer). The output of each recurrent unit $R_l$ is also sent back to the recurrent unit of the previous layer ($R_{l-1}$), providing a top-down signal that allows lower layers to update their predictions based on feedback from higher layers. The PredNet architecture consists of a simple stack of 3 of these recurrent computational units (see Fig. \ref{fig:prednet}B). As is common in deep learning architectures, a max-pooling /sub-sampling operation is applied to the feed-forward error signal, to reduce the spatial dimensions (from 160x128 down to 20x16), along with a corresponding increase in feature (channel) dimension, from 3 RGB channels of the input channels, to 48, 96 and 192 features for each of the recurrent LSTM units, see \cite{lotter_deep_2017} for details). The network is trained by minimizing the output of the prediction error units. Note that even with a fully-trained network, we do not expect the prediction error units to be completely zero (unless the input consists of static frames). Non-zero prediction error is a normal part of network operation. A well-trained network, however, will have a smaller prediction error, since having a better representation of the world states allows each recurrent unit to make more accurate predictions of its inputs. Using this simple objective of minimizing prediction error, PredNet was able to learn to predict the next frame in a sequence of real-world images, taken from the KITTI dataset. 

In the PredNet model, \citet{lotter_deep_2017} showed how upper layers of the network were able to learn more abstract, higher-level representation of the input images. For example, when trained on a dataset of synthetic sequence of rotating faces, the upper layer recurrent units were found to encode latent variables such as rotation angles (roll, tilt) of the faces and face identity better than the lower recurrent units. When trained on KITTI driving sequences, information in the recurrent units was found to correlate with car steering angle. This provided a proof-of-concept that the recurrent representation was learning a more abstract representation of the inputs that could potentially generalize to other tasks. However, the reading out of these latent variables was very sometimes not very competitive with SOTA (e.g. estimation of steering angle had MSE in $\mathrm{degrees}^2$ of 2.14.). This is presumably because the extraction of these latent variables was done post-training. Thus, the embedded information about these implicit state variables may have been entangled with other dimensions of the inputs, and was not optimized for easy read-out using SVMs (the approach used by Lotter et al.)

\paragraph{PreludeNet: combining next-frame prediction and depth inference.} We tested whether the predictive coding architecture could be used to disentangle these implicit state variables and to predict “world-state” information, such as the 3D structure of the visual scene encoded in a depth map, by training the network to be optimized to generate this depth map as part of the training procedure. We selected depth, since this is a highly information-rich domain, and directly useful for many real-world applications such as self-driving cars. Additionally, the internal representation developed by PredNet to predict the next frame, such as optical flow and other motion cues, may be useful for depth estimation, (depth-from-motion). We modified the original PredNet architecture by adding more predictive coding blocks, which successively reversed the feature expansion and spatial reduction, to return to a single-channel depth prediction of the original image dimensions at the top-most layer. We used a wider image input of 418x128 pixels to better match the aspect ratio of the full-image KITTI images, so that the corresponding depth maps would cover the full extent of the ground truth depth maps. (In contrast, the original PredNet architecture was not interested in estimating depth maps and comparing with other depth-prediction benchmarks and so considered only a central 160x128 segment of the down-sampled KITTI images), but otherwise kept the number of feature channels in each of the 3 predictive coding modules the same (48, 96, 192). In addition, we also feed the prediction error at the earlier levels to the corresponding up-sampled upper layers to facilitate incorporation of fine-detailed spatial information that may be lost in the down-sampling layers. Thus, the depth modules are receiving prediction error signals from the corresponding RGB modules and are (re-)interpreting them as depth. 
Note that the symmetry between the two stacks of predictive coding blocks is thus broken in this respect: the RGB-prediction stack provides prediction-error signals to the depth-prediction stack but not the other way around. This is because the predicted RGB frames are being continuously observed and compared with actual next frames, but no ground-truth depth maps are being observed by the network during run-time. 
The resulting U-shaped architecture is similar to (and inspired by) many other depth-prediction networks \cite{zhou_unsupervised_2017, mahjourian_unsupervised_2018, casser_depth_2018}. The resulting architecture is a PRedictive Lateral bottom-Up and top-Down Encoder-decoder Network (PreludeNet), where the "Lateral" refers to the skip connections that connect RGB and depth prediction blocks of the same resolution.

The output of the prediction of the top computational unit is the depth map. In contrast to the predictive unit at the bottom, to which we feed the RGB frames as input to the $A_0$ layer, we do not provide the network with the corresponding ground truth depth during training, since this is data we assume we will not have access to at test time, and we want the network to be able to operate when ground truth depth is not available. Instead, we train the top-most layer by simply adding a supervised depth loss term (absolute difference between the prediction of the $A_5$ unit with ground truth depth) to the loss function. During training, we also add an RGB-gated depth smoothness loss term as in previous works (e.g. \cite{casser_depth_2018}). We could, in principle, replace the supervised loss with a self-supervised training loss as used in the other self-supervised depth prediction networks (\cite{zhou_unsupervised_2017, mahjourian_unsupervised_2018, casser_depth_2018}), but this would require the model to additionally predict camera self-motion output (which is used in those networks, together with the depth map, to jointly learn self motion and depth along with differential frame warping to enforce consistency with the next observed video frame), so we leave this for future work. An additional benefit of using self-supervised depth would be that the depth-prediction stack could also generate its own prediction error, by comparing the observed RGB frame with a predicted RGB frame (using the previous RGB frame along with depth, self-motion and differential frame warping, as mentioned above), which would allow us to restore the symmetry of the connections between the two predictive coding stacks.

\begin{figure}
 \centering
 \includegraphics[scale=0.66]{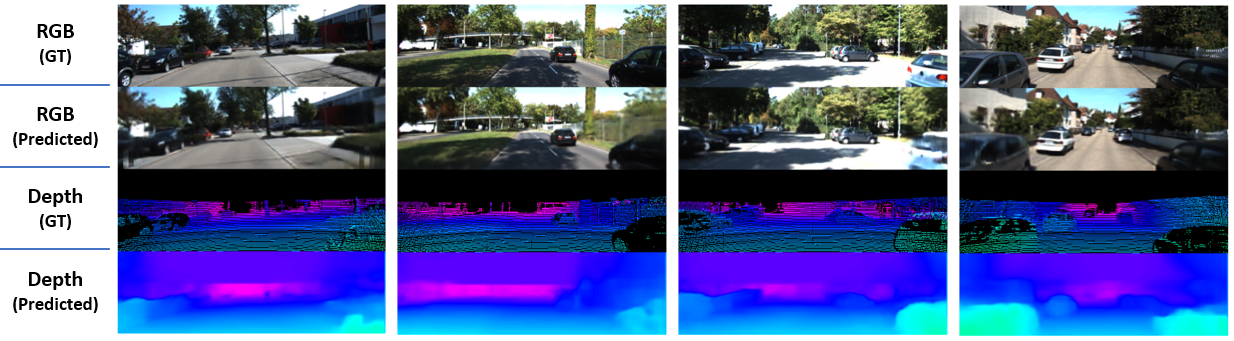} 
 \caption{Examples of predicted depth. Top row: Original RGB frame. Second row: Predicted RGB frame. Third Row: Ground truth depth (LIDAR). Bottom row: Predicted Depth.}
 \label{fig:sample_PreludeNetoutputs}
\end{figure}

\paragraph{Training.} As with the original PredNet model, we trained the network for 150 epochs of 250 iterations each. Each "iteration" consists of taking a random selection of 10 consecutive frames from the training split of the KITTI dataset and feeding each RGB frame sequentially to the network. The procedure for each frame consists of first: (1) Propagating the top-down predictions from the recurrent $R_l$ units, starting from the top-layer and ending at the bottom-most $R_0$ layer. (2) Calculating the $\hat{A}_l$ predictions, and thus prediction error $E_l$ terms for each layer (for more details, see \cite{lotter_deep_2017}). We trained on the KITTI 2015 dataset using the Eigen split \cite{eigen_depth_2014}, and evaluated on the held-out test set. Sample outputs after training the network are show in Figure \ref{fig:sample_PreludeNetoutputs}. 

\section{Results}

\paragraph{Next frame prediction}
We compare the accuracy of the next-frame prediction of PreludeNet with that of PredNet, which has the same architecture but lacks the depth prediction modules. As mentioned above, while the results from the original PredNet study used narrow subsections of the KITTI images (160x128), we wanted to use a model which could also predict full-frame KITTI depth images, so we used the full-width 418x128. 
We retrained the original PredNet model on the full, wide-frame KITTI images instead of the narrower 160x128 images. Our results are show in Table \ref{table:next_frame_results}. PreludeNet achieves slightly better next frame prediction results, indicating that learning to explicitly predict depth improves its ability to predict the next frame.

\begin{table}
  \small
  \caption{Next frame prediction results for the KITTI dataset. "Model Error" indicates the average RGB pixel error between predicted and actual next frames, after dividing RGB values by 255 so that all pixel values are in the range [0, 1]. The first row, "Copy Last Frame" indicates a naive model that predicts the next frame by simply copying the previous frame. "Improvement" for a model indicates the ratio of its "Model Error" to that of the naive "Copy Last Frame" model.}
  \label{table:next_frame_results}
  \centering
  \begin{tabular}{lllll}
    \toprule
    Method        & Model Error & Improvement & SSIM & PSNR \\
    \midrule
    Copy Last Frame &  0.09163 & - & - & -  \\
    PredNet \cite{lotter_deep_2017} & 0.00585 & 15.676 & 0.853 & 23.71 \\
    PreludeNet (ours) & \bf{0.00565} & \bf{16.207} & \bf{0.871} & \bf{23.99} \\
    \bottomrule
    
  \end{tabular}
  
\end{table}

\paragraph{Depth-prediction test.}
The held-out test set from the Eigen split consists of 697 frames, taken from 28 separate driving sequences, by extracting 25 frames sampled approximately evenly from each driving sequence. One complication that arises from our approach is that PreludeNet requires a sequence of consecutive frames, and predictions (of both RGB and depth) are often poor for the first 2-3 frames until the recurrent units “engage” and begin to produce accurate RGB and depth predictions. We thus typically discard predictions for the set of test images that are within 3 frames of the start of a sequence when measuring performance (except in our “static frames” test, below), making it difficult to evaluate on the 31 out of the 697 frames at the beginning of the test sequences (the first frame in each of the 28 sequences, and 3 other frames that were close to the start of very short sequences)  To verify whether leaving out these 31 frames changed our results, we reproduced the results of other methods when including vs excluding these 31 frames. In all cases, results changed by less than 2\%. We thus report the results from our method on the remaining 697-31=666 frames for which PreludeNet can produce accurate depth maps. The depth maps predicted by PreludeNet are have a resolution of 418x128, the same as the input RGB images. Evaluation is done with the full-resolution GT depth maps by using bilinear interpolation to upsample the predicted depth map to the same resolution (typically 1272x375) as the GT depth map.

Results evaluated on the remaining 666 frames are shown in Table \ref{table:depth_results}. Our method well surpasses the initial self-supervised baseline set by \citet{zhou_unsupervised_2017}, as well as the competitive struct2depth model \cite{casser_depth_2018}, which uses more advanced motion masks. 

\paragraph{Depth-delayed test.}
It is important to note that in these tests just described, our network is actually performing \emph{next-frame depth prediction}, i.e. predicting the depth map for an RGB frame which has not yet been seen. This is because at time step 
$t_i$, the recurrent units make their predictions (for both the RGB and depth outputs) based only on the RGB inputs from the previous time step $t_{i-1}$.  We can make the task slightly easier by making information about the RGB frame at time $t_i$  available when it is making the depth prediction from time $t_i$. This can be achieved by modifying the training and simply delaying the provided sequence of ground truth depth images by one frame, such that the network is now trained, at each time-step $t_i$, to predict the RGB frame from $t_i$ and the depth frame from time 
$t_{i-1}$.  Introducing this depth-delay resulting in a modest improvement (decrease in depth error metrics by up to 6\% ). It is worth noting that results for this depth-delayed task is more comparable to that of previous works, since the networks is estimating depth for a frame that has already been observed). In contrast, the depth-\emph{prediction} task is a somewhat more difficult task, since the network is predicting the depth map for an RGB frame that has not yet been observed. The fact that the network has only slightly worse performance for the depth prediction task points to the robustness of the PreludeNet model.

\paragraph{Static Frames test.} Our intuition was that PreludeNet learns to predict depth by extracting latent information from scene motion, i.e.. the relative speed at which components of the image are moving across the optical field, and not just learning to do an RGB->depth “translation” that does not take into account any motion cues. We tested this hypothesis by training and testing PreludeNet with only static sequences (i.e. the sequence of 10 frames is just the same frame repeated 10 times). As shown in Table 1 (static frames test), the results are slightly worse compared to when using consecutive frames, indicating that some cues from motion are indeed being exploited for depth prediction. The effect size is small, however (an increase in depth error metrics by up to only 2\%). Thus, the use of recurrent units allows the network to take some advantage of motion cues, but this seems to provide only a modest increase in performance. 

\paragraph{Multiple frames test.} The recurrent LSTM modules in PreludeNet generate an internal model of how the RGB input changes from frame to frame, and extrapolating that change to predict the next frame. We wondered whether the model would benefit from being explicitly provided with more than one frame from the video history when predicting the next frame. To test this, we modified the PreludeNet architecture to accept multiple frames, while still predicting one frame in advance. For example, using 3 frames a time, at t=0, the network is given frames 1,2,3, and is trained to predict frames 2,3,4. Since the network can easily learn to just copy the overlapping frames, the loss function (and evaluation metrics) consider accuracy for only the newest (unseen) frame, (i.e., frame 4 in this example). Similarly, for consistency, while a depth map is produced for each corresponding RGB frame, only the depth map corresponding to the newest image is used for the loss function and in calculating the depth accuracy metrics. Results for using 2 or 3 frames are shown in Table \ref{table:depth_results}. In both cases, results were the same or slightly worse compare to using a single frame, indicating that no advantage is gained by using multiple frames.

\begin{table}
  \small
  \caption{ Single-view depth results on the KITTI dataset \protect\cite{geiger_are_2012}  using the split of Eigen \protect\cite{eigen_depth_2014}  (Baseline numbers taken from \protect\cite{godard_unsupervised_2017}). Entries with a * indicate methods that use supervised learning with ground-truth depth }
  \label{table:depth_results}
  \centering
  \begin{tabular}{lllllllll}
    \toprule
    Method        & Abs Rel & Sq Rel & RMSE & RMSE log & $\delta < 1.25$ & $\delta < 1.25^2$ & $\delta < 1.25^3$ \\
    \midrule
    Training Set Mean &  0.361 & 4.826 & 8.102 & 0.377 & 0.638 & 0.804 & 0.894 \\
    \citet{eigen_depth_2014}*  & 0.214 & 1.605 & 6.563 & 0.292 & 0.673 & 0.884 & 0.957 \\
    \citet{zhou_unsupervised_2017} & 0.208 & 1.768 & 6.856 & 0.283 & 0.678 & 0.885 & 0.957 \\
    \citet{yang_lego_2018} & 0.182 & 1.481 & 6.501 & 0.267 & 0.725 & 0.906 & 0.963 \\
    \citet{mahjourian_unsupervised_2018} & 0.163 & 1.240 & 6.220 & 0.250 & 0.762 & 0.916 & 0.968 \\
    \citet{yang_lego_2018} & 0.162 & 1.352 & 6.276 & 0.252 & 0.783 & 0.921 & 0.969 \\
    \citet{yin_geonet_2018} & 0.155 & 1.296 & 5.857 & 0.233 & 0.793 & 0.931 & 0.973 \\
    \citet{wang_learning_2018}  & 0.151 & 1.257 & 5.583 & 0.228 & 0.810 & 0.936 & 0.974 \\
    \citet{godard_unsupervised_2017} & 0.133 & 1.158 & 5.370 & 0.208 & 0.841 & 0.949 & 0.978 \\
    \citet{casser_depth_2018} ('M') & 0.141 & 1.026 & 5.291 & 0.215 & 0.816 & 0.945 & 0.979 \\
    \citet{yang_every_2018} & 0.137 & 1.326 & 6.232 & 0.224 & 0.806 & 0.927 & 0.973 \\
    \citet{yang_every_2018}  & 0.131 & 1.254 & 6.117 & 0.220 & 0.826 & 0.931 & 0.973 \\
    \citet{bian2021unsupervised} & \bf{0.114} & \bf{0.813} & \bf{4.706} & \bf{0.191} & \bf{0.873} & \bf{0.960} & \bf{0.982} \\
    \midrule 
    Ours* [standard] & 0.138 & 0.973 & 5.024 & 0.203 & 0.822 & 0.945 & 0.979 \\
    Ours* [depth delayed] & 0.131 & 0.911 & 5.179 & 0.200 & 0.831 & 0.945 & 0.980 \\
    Ours* [static frames] & 0.141 & 0.951 & 5.154 & 0.208 & 0.822 & 0.942 & 0.979 \\

    Ours* [standard; 2 frames] & 0.139 & 0.998 & 5.061 & 0.204 & 0.822 & 0.941 & 0.978 \\
    Ours* [standard; 3 frames] & 0.141 & 1.035 & 5.085 & 0.205 & 0.823 & 0.939 & 0.978 \\

    \bottomrule
    
  \end{tabular}
  
\end{table}

\subsection{Illumination invariance tests for depth inference}

Since one of our goals was to evaluate the predictive coding computational motif as an effective method to produce reliable, robust estimates of `world state', we wanted to directly evaluate the robustness of our method to variations in appearance and lighting conditions. Algorithms that are more robust to such variations will fare better in real world environments where variations in lighting and shadows can drastically alter appearance. There are several publicly available synthetic datasets (e.g. Virtual KITTI \cite{gaidon_virtual_2016}) that contain variations in lighting conditions, but not with the fine-grained control that we were looking for. We thus generated a new Synthetic Invariant NAVigation video (SINAV) dataset, generated using a custom SINAV app based on the Unity3D Engine \cite{unity_unity_2017}. The SINAV app is designed to produce synthetic navigation datasets with controlled variations in image properties, allowing us to parametrically vary lighting conditions while keeping other variables constant. Similar to other navigation datasets such as KITTI, our dataset contains images captured from the perspective of an autonomously moving agent. The agent explores the virtual environment, while the SINAV app generates RGB, depth, optical flow, and semantic segmentation images with a virtual camera. In this work, we have only used the RGB and depth channels, but this app was designed with future models which may take advantage of more of these modalities. 

The environment is set up with a variety of natural objects (rocks and trees) of various dimensions, some vehicles, and geometric shapes (cuboids, spheroids, and cylinders) on top of a flat terrain. Illumination is controlled by combination of sun (a source of  directional light, but is not otherwise visible) and ambient light. The SINAV app is able to generate various illumination effects, such as simulating different times of the day (morning, afternoon, sunset, night, etc.) with different exposure, intensity and color of sun light and ambient light, and direction of the sun light.

The autonomous agent navigates through the environment along a smooth trajectory with varying speed and while avoiding collision with other objects (using an algorithm implemented using RVO2 \cite{noauthor_reciprocal_nodate}). The agent has a headlight with adjustable intensity and range, which can be another light source when switched on. Camera angle and direction can be varied independently of the robot motion thus allowing sample trajectories with a high variety in optical flow.   

\begin{figure}
  \centering
  \includegraphics[scale=0.80]{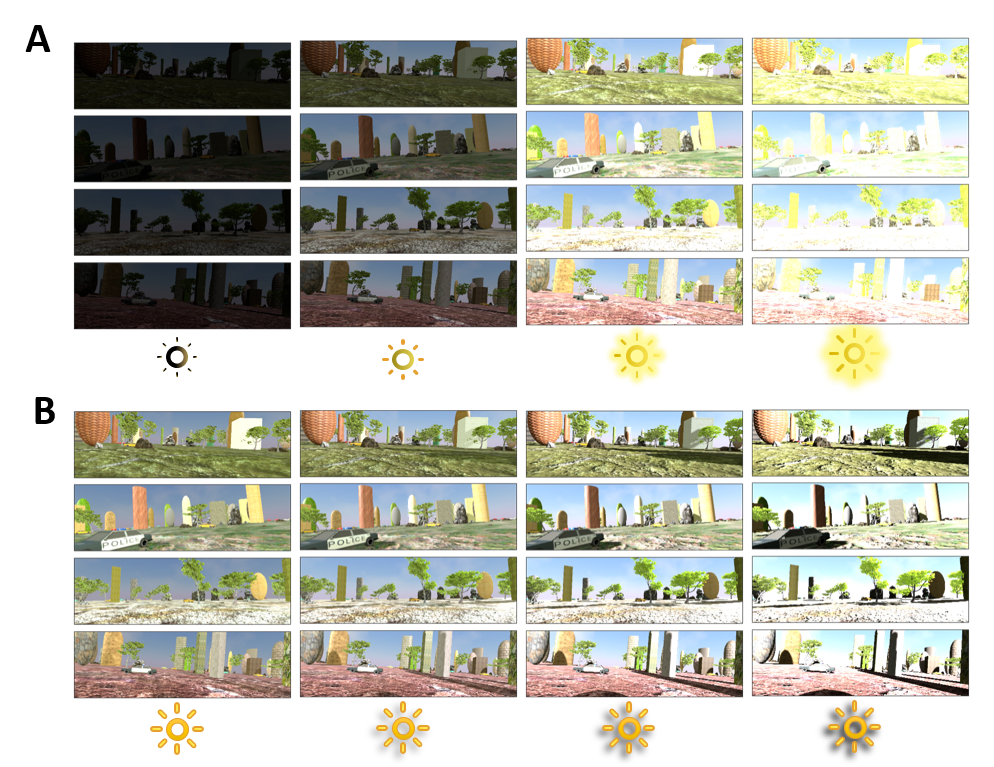}  
  \caption{ Samples from our SINAV (Synthetic Invariant NAVigation) dataset. \textbf{A}. Samples from our “Illumination” test dataset. Each row contains a snapshot from a one of the synthetic worlds created using our SINAV app, under varying lighting conditions from dim to bright \textbf{B}. Samples from our “Shadows” test dataset, in which the ratio of ambient to direct sunlight is varied, to control the brightness and contrast of shadows.}
 \label{fig:sinav_samples}
\end{figure}

We generated synthetic scenes to perform two tests. (1) an “Illumination” test, to measure robustness to changes in total amount of light, from very dim (e.g. pre-dawn) to highly over-saturated (bright midday) lighting conditions. In this test, both sun and ambient lighting are increased and decreased together on an arbitrary scale from 1 to 10 (see examples in Fig. \ref{fig:sinav_samples}A). (2) A “Shadows” Test, to measure robustness to differences in appearance between diffuse lighting with few shadows and strong sunlight with high-contrast shadows. In this test, we adjust the ratio of sunlight to ambient light on an arbitrary scale from 1 to 10. (Fig. \ref{fig:sinav_samples}B).  

\begin{figure}
  \centering
  \includegraphics[scale=0.65]{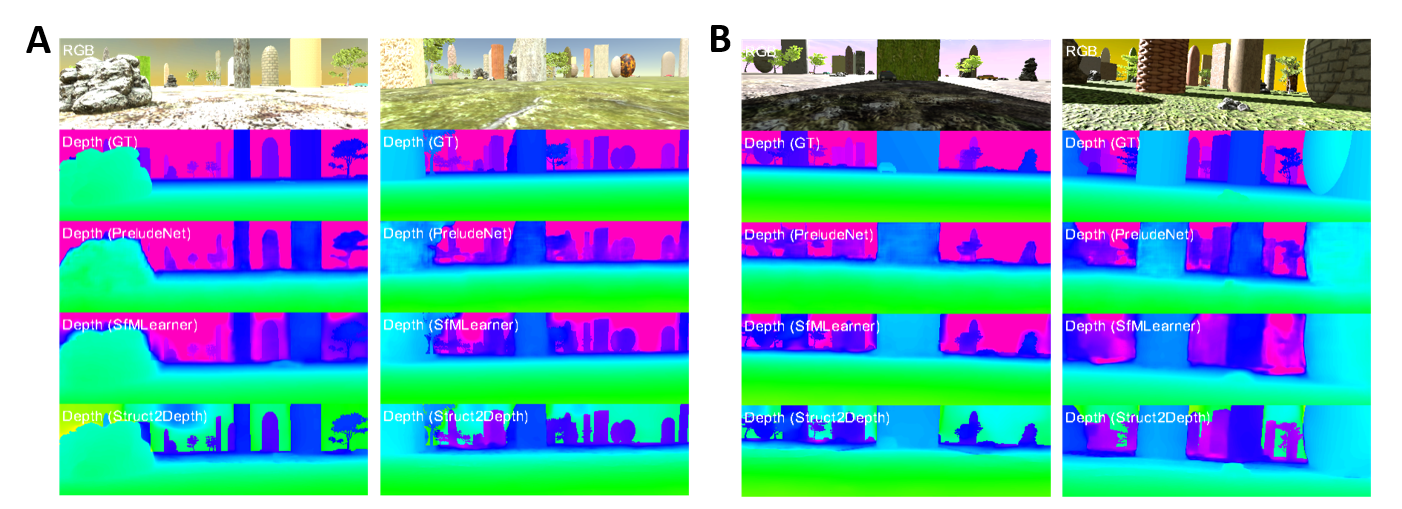}  
  \caption{ Invariance test output samples, for illumination test (A) and shadows test (B). Rows (from top to bottom) show: (1) RGB; (2) Ground-truth depth; (3) PreludeNet depth (4) SfMLearner depth and (5) struct2depth depth.}
 \label{fig:sinav_predictions}
\end{figure}

\begin{figure}
  \centering
  \includegraphics[scale=0.6]{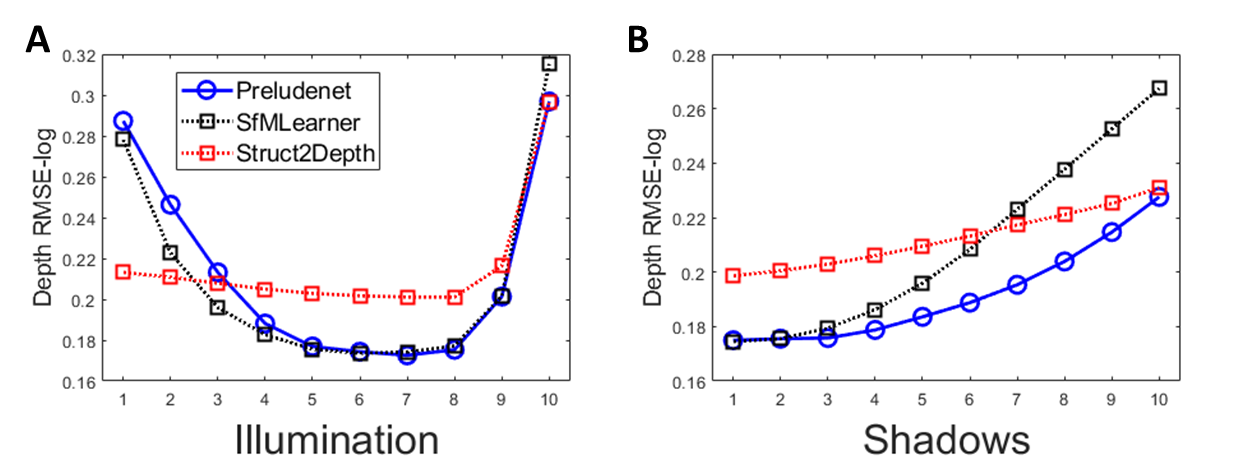}  
  \caption{ Results of Invariance tests. A. Results of Illumination test. Sample depth error metric (Depth RMSE-log) vs illumination for PreludeNet, SfmLearner and struct2depth. B. Similar to A, but for Shadows test.}
 \label{fig:sinav_plots}
\end{figure}

For each of the two lighting tests, we generated a “training set”, which consists of 1000 random worlds, with a random trajectory consisting of 50 frames in each world, and with lighting conditions sampled randomly from a subset of our predefined (arbitrary) lighting scale (e.g. lighting scale 3 to 8), excluding the most extreme examples on either end of the lighting scale (e.g. 1,2 and 9,10); (2) a test set, in which 100 new worlds (distinct from those in the training set) are each repeatedly simulated for each of the 10 values along the lighting scale.\footnote{To facilitate future research, these training and testing datasets will be made public after publication.} To compare robustness to other methods, we compared against SfMLearner\cite{zhou_unsupervised_2017} and the more sophisticated struct2depth  \cite{casser_depth_2018} algorithms. Curiously, both of these approaches were unable to learn to predict depth when directly trained on the SINAV dataset alone, for reasons which remain unclear. However, if first pretrained on the KITTI dataset, they could be successfully fine-tuned on the SINAV dataset. (In contrast, PreludeNet was able to directly learn to predict depth on the SINAV dataset without pretraining on KITTI). However, for the sake of a fair comparison, we used the same training regime for PreludeNet, first training on KITTI and then finetuning on SINAV. Sample outputs from each of these networks (as well as the ground truth depth) are shown in Figure \ref{fig:sinav_predictions}. Qualitative analyses of results are shown in Figure \ref{fig:sinav_plots}.

For the illumination test, results are inconclusive. PreludeNet and SfMLearner achieve similar depth metrics across the series. While for intermediate illuminations, their depth estimates are superior to those from struct2depth, the quality of their estimates degrades on the more extreme ends of the lighting scale. In contrast, struct2depth has higher depth error for intermediate illumination levels, but degrades more gracefully at lower absolute lighting. Results are more promising for the Shadows test, however. PreludeNet achieves better depth estimates than both algorithms even under high contrast conditions. These tests highlight the importance of measuring robustness to different conditions. We plan to release the training/testing dataset for the Illumination Invariance tests as they may be a useful to other researchers as a benchmark for testing model invariance. As mentioned above, there are other publicly available synthetic navigation datasets, but not with fine control over various lighting/shadow conditions as shown here.

\section{Conclusion}

Predictive coding can be an effective framework for representing world state parameters such as depth, achieving close to state-of-the-art accuracy in depth on the KITTI dataset, and competitive accuracy on SINAV invariance tests as compared to the models with implicit 3D inductive bias on the Shadows test. While the accuracy of depth maps is not that far beyond the state-of-the art, our contribution in this paper is to demonstrate a novel approach, using a predictive coding framework to extract information about world state parameters and evaluate their quality in terms of relevant invariances. 
One of the advantages of using predictive coding (i.e. as opposed to the standard feed-forward connections of typical deep networks) is the opportunity to train using self-supervision: learning a representation of world states by learning to predict future input frames. In this work, we showed that, using the predictive coding computational motif, we can learn world states from which we can extract accurate depth maps that are competitive with state-of-the art, but which still use direct depth supervision during training. 
In future work, we plan to evaluate the contribution of depth supervision to use hybrid and self-supervised techniques to predict multiple parameters of the ‘world states’, including self-motion, optical flow and semantic segmentation.

\section*{Broader Impact}

This work highlights the potential advantages of using a predictive coding in deep learning frameworks, and shows how the predictive coding computational motif can be used not only for video extrapolation (predicting the next frame), but also to extract useful implicit information about world states from the learned feature representation. In addition, our results suggest that incorporating the principles of predictive coding into future deep learning frameworks may increase robustness and reliability of autonomous agents.




\bibliographystyle{plainnat}
\bibliography{truevis_neurips}

\end{document}